\documentclass[journal]{IEEEtran}
\ifCLASSINFOpdf
\else
   \usepackage[dvips]{graphicx}
\fi
\usepackage{amsmath}
\usepackage{array}

\ifCLASSOPTIONcompsoc
  \usepackage[caption=false,font=normalsize,labelfont=sf,textfont=sf]{subfig}
\else
  \usepackage[caption=false,font=footnotesize]{subfig}
\fi
\usepackage{url}
\usepackage[colorlinks=true]{hyperref}
\usepackage[hyphenbreaks]{breakurl}

\usepackage{mathptmx}
\usepackage{graphicx}
\usepackage{times}
\usepackage{paralist}

\usepackage{amsmath,amssymb}
\usepackage{color}
\usepackage{xspace}
\usepackage[table,xcdraw]{xcolor}
\usepackage[normalem]{ulem}
\usepackage{bm}

\usepackage[demo,percent]{overpic}

\makeatletter
\DeclareRobustCommand\onedot{\futurelet\@let@token\@onedot}
\def\@onedot{\ifx\@let@token.\else.\null\fi\xspace}
\def\etal{~et~al\onedot}
\def\eg{e.g\onedot} 
\def\ie{i.e\onedot}

\makeatother

\def \RR {{\mathbb R}}

\newcommand{\norm}[1]{ \left\| #1 \right\| }

\begin{document}

\title{Residuum-Condition Diagram and Reduction of Over-Complete Endmember-Sets}

\author{
  \IEEEauthorblockN{Christoph~Schikora, Markus Plack, and~Andreas~Kolb}
  \thanks{Authors are with the Computer Graphics and Multimedia Systems Group, University of Siegen, 57076 Siegen, Germany}
}



\maketitle

\begin{abstract}
Extracting reference spectra, or endmembers (EMs) from a given multi- or hyperspectral image, as well as estimating the size of the EM set, plays an important role in multispectral image processing. In this paper, we present \emph{condition-residuum-diagrams}. By plotting the residuum resulting from the unmixing and reconstruction and the condition number of various EM sets, the resulting diagram provides insight into the behavior of the spectral unmixing under a varying amount of endmembers (EMs). Furthermore, we utilize condition-residuum-diagrams to realize an \emph{EM reduction algorithm} that starts with an initially extracted, over-complete EM set. An over-complete EM set commonly exhibits a good unmixing result, \ie a lower reconstruction residuum, but due to its partial redundancy, the unmixing gets numerically unstable, \ie the unmixed abundances values are less reliable. Our greedy reduction scheme improves the EM set by reducing the condition number, \ie enhancing the set's stability, while keeping the reconstruction error as low as possible. The resulting set sequence gives hint to the optimal EM set and its size. We demonstrate the benefit of our condition-residuum-diagram and reduction scheme on well-studied datasets with known reference EM set sizes for several well-known EE algorithms.
\end{abstract}


%
\IEEEpeerreviewmaketitle

\section{Introduction}
\label{intro}
\IEEEPARstart{T}{he} estimation and detection of constituting materials, \ie \emph{end members (EMs)} in multi- or hyperspectral imagery (we will use multispectral as synonym for both), and the unmixing of the given dataset with respect to extracted EMs, is an important step for classification and structural analysis in fields such as remote sensing and spectral microscopy. In \emph{spectral mixture analysis} a linear mixture model is assumed and the spectral unmixing with respect to the constituent EMs provides their abundances at a per-pixel level representing the material fractions~\cite{BioucasDias2012,Somers2011}. Commonly, \emph{full-constrained} linear unmixing is applied, yielding   non-negative abundance values that sum up to one.

Endmember extraction (EE) algorithms extract EMs from the multispectral image directly~\cite{Somers2011}. Without a-priori information, EE algorithms need to determine a minimal set of ``pure'' endmembers at acceptable computational costs, whose linear unmixing results in a proper reconstruction of the initial spectral image~\cite{Sabol1992}. The size of the EM set can be estimated by simultaneous extraction and unmixing approaches (guided by manual or statistic thresholds) or by automatic data driven decisions, such as virtual dimensionality methods~\cite{Asl2016}.

In this paper, we propose \emph{condition-residuum diagrams} that relate the EM matrix's condition number $\kappa$ and the root mean square error (RMSE) as residuum measure of the reconstruction after unmixing. Our approach is inspired by the observation that larger EM sets lead to a better reconstruction after unmixing, but at the cost of spectral redundancy that, on the down-side, makes unmixing, and thus the reconstruction result numerically unstable~\cite{Sabol1992}. Condition-residuum diagrams provide deeper insight into the relation between redundancy (or instability) and the residuum after unmixing for various EM sets for a given multispectral image. Based on condition-residuum diagrams, we propose an \emph{EM reduction algorithm} that is applied to a given, over-complete EM set in order to semi-automatically identify the ``best'' subset of EMs. Here, ``best'' means that the desired EMs set exhibits a \emph{low residual error} (after unmixing and reconstruction) and a \emph{low condition number} (indicating numerical stability). Our greedy EM reduction approach determines a nested  sequence of EM subsets yielding maximum stability at minimal residuals. Together, the condition-residuum diagram and the reduction algorithm provide quantified means of selecting a proper EM set and insight into the general composition of the multispectral image with respect to the unambiguity of its endmembers.

We evaluate our procedure using different multispectral datasets with three EE algorithms, showing the usability of our visual condition-residuum diagram and our EM reduction scheme with respect to the quality of the deduced EM sets.

\section{Related Work}
\label{related}
Several direct EE algorithms, that do not involve an explicit unmixing have been developed~\cite{Plaza2004}. The Pixel Purity Index (PPI) of Boardman\etal\cite{Boardman1993} projects spectra from the dataset onto randomly selected vectors in order to find vertices of a convex hull of the multispectral data. Orthogonal Subspace Projection (OSP) by Harsanyi and Chang~\cite{Harsanyi1994} recursively selects the maximum projection of the spectra in the subspace orthogonal to the span of the current EM set. The N-FINDR algorithm of Winter~\cite{Winter1999} is a simplex growing approach that selects and refines the EM set by maximizing the simplex's volume. Similar to OSP, the Vertex Component Analysis (VCA) algorithm uses a subspace projection scheme, but generates an intermediate simplex that is used to identify the EMs via projection~\cite{Nascimento2005}. The Iterative Error Analysis (IEA) of Neville\etal\cite{Neville1999} is an iterative EE process that selects the pixel (or an averaged pixel set) within the image as new EM that exhibits the maximal residuum after unmixing.

Other approaches iteratively optimize EM sets using spectral unmixing. Based on a direct (iterative) EE algorithm, they use manual residuum thresholds, in-/stability thresholds, or data driven, statistical thresholds, to optimize the EM set. Van~der~Meer~\cite{VanDerMeer1999} presents an iterative spectral unmixing approach optimizing the EM set generated by PPI, by iteratively exchanging EMs according to the residuals error in their pixel neighborhood. Song\etal\cite{Song2015} present an EM optimization based on IEA, which excludes EMs with a low residuum gain in their IEA order and EMs with a small spectral angle to the first three EMs. 

Plaza and Chang~\cite{Plaza2006} investigate the influence of termination rules applied to EE algorithm with respect to the EM quality. They demonstrate that if the number of extracted EMs is too small, relevant spectra are not extracted and when the number is too high, interfering substances, \ie, very similar spectra are selected. 

Berman\etal\cite{Berman2004} introduced the statistical iterated constrained endmember (ICE) algorithm. This approach solves all tasks in parallel, \ie endmember selection, unmixing and the determination of the number of endmembers, by combining statistical analysis with the attempt to optimally cover the simplex formed by the scene pixels' spectra. Zare and Gader~\cite{Zare2007} extend the ICE algorithm by adding a sparsity promotion scheme. Both approaches generate ``synthetic'' endmembers that are in most cases not in the given data. In contrast, our method focus the selection of endmembers that are explicitly given in the data to be analyzed.

In general, even if the ``correct'' EM set size is known, both, EE algorithm and EM set optimization approaches, often do not extract all relevant EMs. To the best of our knowledge, no EE or optimization algorithm delivers a reliability measure that involves both, reconstruction quality (RMSE) and unmixing stability. This approach, as shown in this paper, is less sensitive to initial parameter setting (we have only one parameter, that we fixed) provided that the over-complete set is large enough. 

\begin{figure*}[!t]
\centering
\begin{tabular}{cccc}
\rotatebox[origin=c]{90}{\hspace*{24mm} Salinas-A} &
\subfloat[OSP]{\includegraphics[width=0.26\textwidth,trim=0 14 0 8,clip]{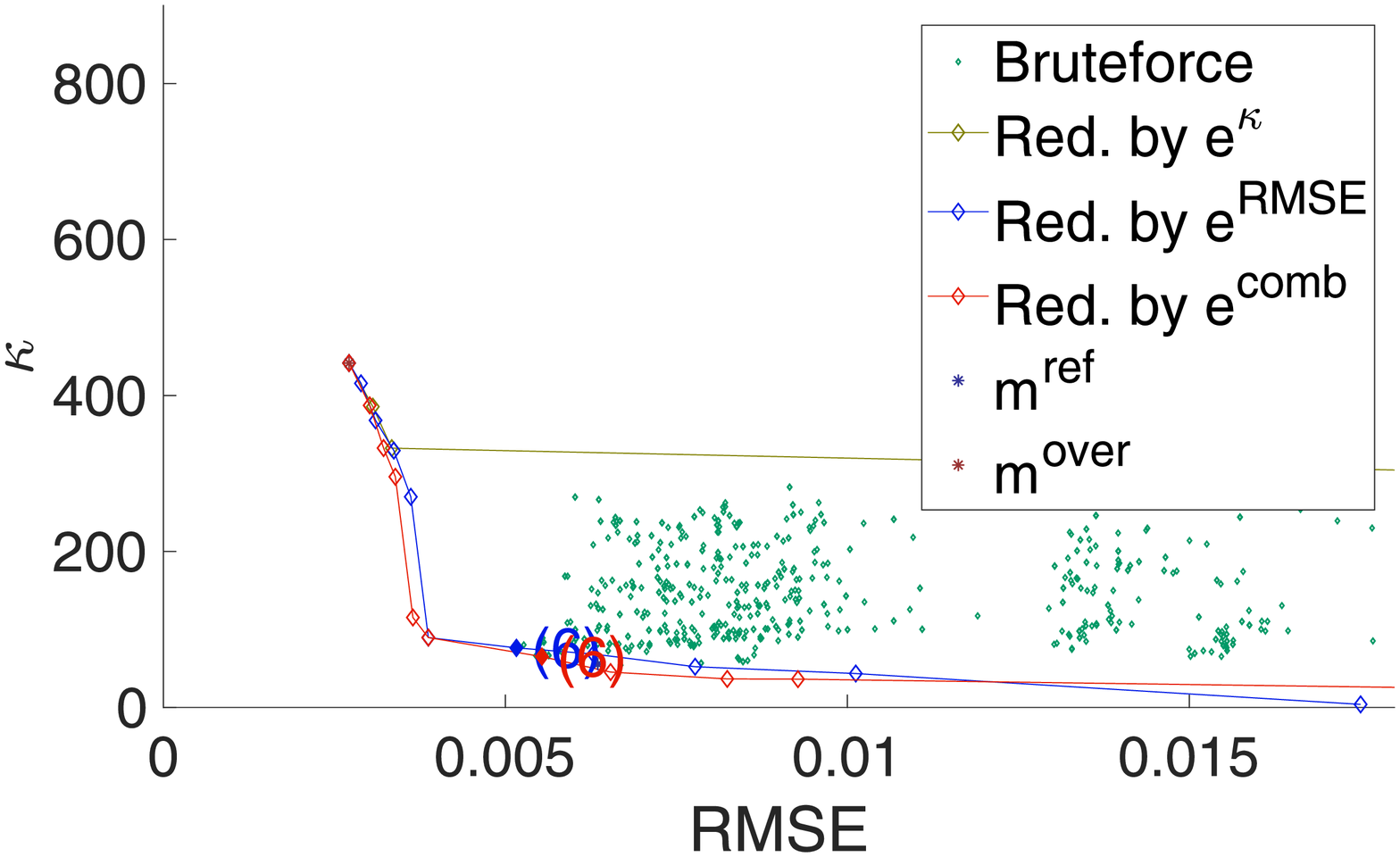}
\label{fig_1a}} &
\subfloat[N-FINDR]{\includegraphics[width=0.26\textwidth,trim=0 14 0 8,clip]{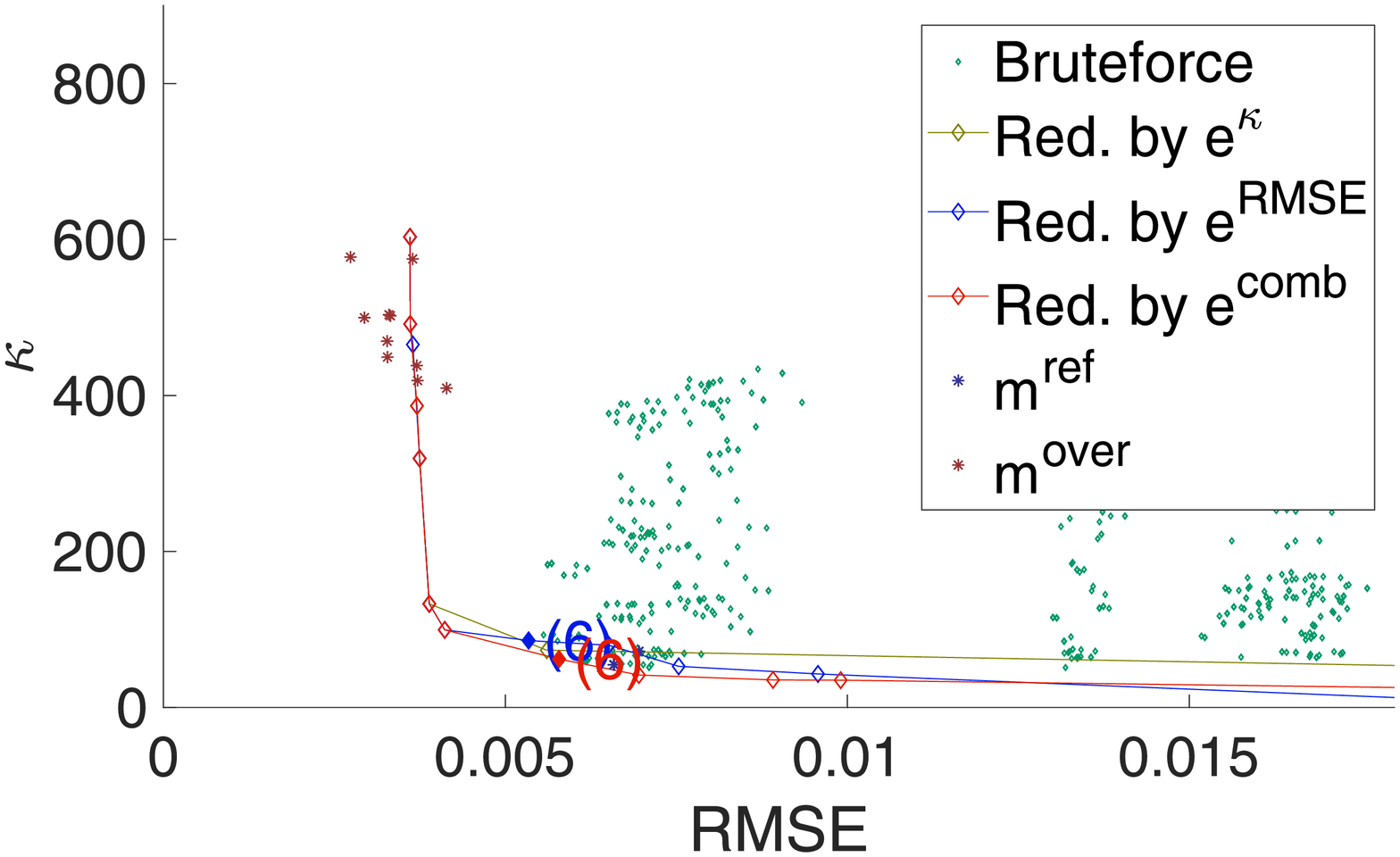}
\label{fig_1b}} &
\subfloat[VCA]{\includegraphics[width=0.26\textwidth,trim=0 14 0 8,clip]{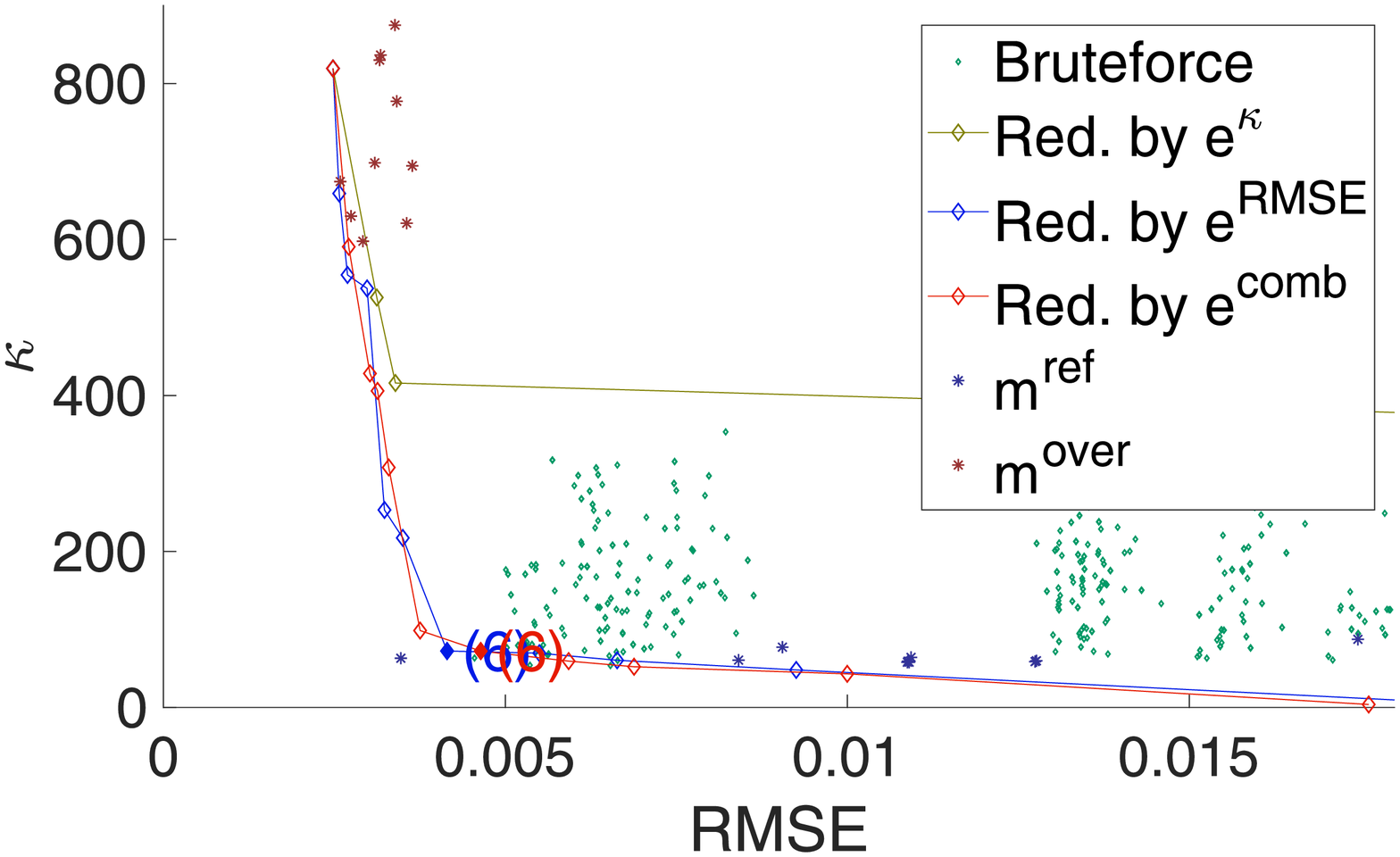}
\label{fig_1c}}
\end{tabular}\\[-4em] 
\caption{Results, Salinas-A: \eqref{fig_1a} OSP, \eqref{fig_1b} N-FINDR, \eqref{fig_1c} VCA.}
\label{fig_1}
\end{figure*}

\section{Method}
\label{method}
Our method comprises of two items. The condition-residuum diagram that provides insight into the relation between the stability of spectral unmixing and the unmixing residuum is described in Sec. \ref{methodA}. In Sec. \ref{methodB} we introduce our EM reduction scheme applied to over-complete EM sets.

\subsection{Condition-Residuum-Diagram}
\label{methodA}
Condition-residuum-diagrams visualize the relation between the error measure of the image reconstruction after spectral unmixing and the condition number of the EM set that is a measure for the instability of the unmixing. Given an endmember-set $\mathcal{S}=\{e_1,\ldots,e_m\}$ consisting of $m$ EMs $e_i\in\RR^n$, $E$ is the $n \times m$ EM matrix, in which the EMs are arranged as columns. We choose the \emph{Root Mean Square Error (RMSE)} as \emph{residuum} measure of $\mathcal{S}$ with respect to the underlying multispectral image $I$.
\begin{equation}
  \text{RMSE}(S) = \frac1{\sqrt{n\cdot p}}\norm{(E\cdot A-I)}_{F}.
  \label{eq:rmse}
\end{equation}
Here, $\norm{.}_F$ is the Frobenius norm that is applied to the difference between the image reconstruction using the abundance matrix $A\in\RR^{m\times p}$ resulting from the spectral unmixing, and the multispectral image $I$. $p$ denotes the number of pixel in image $I$.

To measure the (in)stability of a given EM set $\mathcal{S}$ we choose the matrix condition number that measures the stability of the linear transformation given by a matrix, \ie, how much the output value of the linear function can change for a small change in the input argument. According to van~der~Meer and Jia~\cite{Van2012}, the condition number is a direct measure for collinearity of a given EM set. The matrix condition number $\kappa$ of an EM set $\mathcal{S}$ is computed as the ratio between the largest and the smallest singular value of the matrix $E$ composed of the EMs in $\mathcal{S}$.

The \emph{condition-residuum-diagram} plots the condition numbers and residuum values of several EM sets in order to assess their individual numerically instabil in the unmixing process in relation to their resulting reconstruction residual error (see Eq.~\eqref{eq:rmse}) after unmixing. The diagram supports the simultaneous evaluation both quality criteria and, thus, a revealing means for comparing different endmember sets. The ``ideal'' EM sets exhibits a condition number of $1$ and a residuum of $0$. In practice, there are no ideal EM sets, thus an EM set either exhibits a significant residual error, \ie it does not reconstruct the image $I$ very well, or the EM set is partially redundant, which restricts the numerical unmixing stability. Finding a ``good'' EM set can visually be interpreted as finding an EM set close to the ideal point in the condition-residuum-diagram (see discussion in Sec.~\ref{evaluation}). 

\subsection{Reduction of Over-Complete EM Sets}
\label{methodB}
Based on the condition-residuum-diagram, we propose an EM reduction scheme that is applied to an over-complete EM set $\mathcal{S}_m$ and that results in a sequence of nested sets $\mathcal{S}_m,\mathcal{S}_{m-1}=\mathcal{S}_m\setminus\{e_m\},\ldots,\mathcal{S}_{1}=\mathcal{S}_2\setminus\{e_2\}$ by iteratively removing endmembers $e_i$. 

The main idea in selecting an EM $e$ for removal of the current set $\mathcal{S}_i$ is to optimize the remaining set $\mathcal{S}_i\backslash\{e\}$ to have as low as possible residual error (see Eq.~\eqref{eq:rmse}) and condition number. This approach follows the basic principle that an ``optimal'' EM set should describe the spectral variability of the dataset with a minimal EM set size~\cite{Sabol1992}. We solve this multi-critera optimization problem by combining both measures. Thus $\mathcal{S}_{i-1}$ results from $\mathcal{S}_i$ by removing $e_i^\alpha$ given as
\begin{align}
  e_{i}^{\alpha} = \arg\max_{e\in \mathcal{S}_i}&\left((1-\alpha)\left(\frac{\kappa(\mathcal{S}_i)-\kappa(\mathcal{S}_i\backslash\{e\})}{\kappa(\mathcal{S}_i)}\right)\right.\notag\\
& +\left.\alpha\left(\frac{\text{RMSE}(\mathcal{S}_i)-\text{RMSE}(\mathcal{S}_i\backslash\{e\} )}{\text{RMSE}(\mathcal{S}_i)}\right)\right)
  \label{eq:cost-func}
\end{align}
Reducing an EM set naturally results in a descending condition number and in an ascending RMSE. Thus, our optimization approach maximizes the gain in condition number and minimizes the loss in RMSE. Our scheme works on normalized measures as the absolute value in the measures are not comparable. 

In our empirical evaluation we found $\alpha=\frac12$ a good choice for the $\alpha$-parameter. Therefore, we use $\alpha\in\{0,1,\frac12\}$ in Sec.~\ref{evaluation}, which leads to EMs $e_i^{\kappa}, e_i^{\text{RMSE}}$ and $e_i^{\text{comb}}$ selected for reduction that depend purely on the condition number $\kappa$, purely on the RMSE, and equally on both measures, respectively. Considering the combined reduction with $\alpha=\frac12$, our procedure will select $e^{\text{comb}}$ such that the residual error stays small, while the condition number decreases as much as possible, resulting in a more stable EM set. 

We deliberately do not propose an ``optimal'' EM set size, based on our reduction approach. The ``optimal'' sizes of the EM sets used in our evaluation (Sec.~\ref{evaluation}) have slightly varying position in the condition-residuum-diagram, \ie application specific considerations play an important role. Furthermore, even sophisticated automatic EM set size estimators such as HySime~\cite{Bioucas2008} are often far off the reference size (see Tab.~\ref{table_datasets1}).

\section{Evaluation}
\label{evaluation}
We evaluate the condition-residuum-diagrams and our EM reduction scheme using various sample datasets (see Sec. \ref{Datasets}). In Sec.~\ref{E_CCRD} we present the main properties of the diagram and the reduction scheme and compare the different reduction schemes based on Eq.~\eqref{eq:cost-func}, \ie using solely the condition number $\kappa$ or RMSE, or the combined version.

\subsection{Datasets and Endmember Extraction Algorithms}
\label{Datasets}

\begin{table}[!t]
\caption{Datasets}
\label{table_datasets1}
\begin{IEEEeqnarraybox}[\IEEEeqnarraystrutmode\IEEEeqnarraystrutsizeadd{4pt}{2pt}\IEEEsetsidemargin{c}{5pt}][b][\columnwidth]{t+t+t+t+t+t}
\IEEEeqnarraydblrulerowcut\\
  Name & Size & Bands & $m^{\text{ref}}$ & HySime\\
	\hline
	Salinas-A & 86x83 & 204 & 6 & 18\\
	\hline
	Pavia University & 610x340 & 103 & 9 & 60\\
	\hline
	Cuprite & 250x191 & 188 & 12 & 18\\
	\hline
	Kennedy Space Center & 512x614 & 176 & 13 & 2\\
	\hline
	Indian Pines & 145x145 & 200 & 16 & 18\\
\IEEEeqnarrayseprow[4pt]{}\\
\IEEEeqnarraydblrulerowcut
\end{IEEEeqnarraybox}
\end{table}
Table \ref{table_datasets1} depicts the parameters of the dataset used for evaluation. We use datasets for which reference numbers $m^{\text{ref}}$ for the ``best'' EM set size are known in literature (column $m^{\text{ref}}$). We give the EM set sizes as estimated by the HySime algorithm~\cite{Bioucas2008} as further reference. The Cuperite dataset is online available at \cite{Data1} and the other datasets at \cite{Data2}.

We choose the OSP, N-FINDR and VCA as EE algorithms for our evaluation, where we use our own implementation of OSP and the N-FINDR and VCA implementations of the Hyper Spectral Toolbox \cite{HyperSpectralToolBox}. As suggested by Plaza\etal~\cite{Plaza2011}, we deactivated the noise reduction stage of VCA for a fair comparison. Beside this we use the online available implementation \cite{UnmixingAlgo} of constrained least squares unmixing of Chouzenoux\etal\cite{Chouzenoux2014}. For every EE algorithm we compute an over-complete endmember-set with twice the reference size (\ie $m^{\text{over}}=2m^{\text{ref}}$), and reduce it using our greedy reduction algorithm (see Sec.~\ref{methodB}). 

\begin{figure*}[!t]
\centering
\begin{tabular}{cccc}
\rotatebox[origin=c]{90}{\hspace*{24mm} Pavia University} &
\subfloat[OSP]{\includegraphics[width=0.26\textwidth,trim=0 14 0 8,clip]{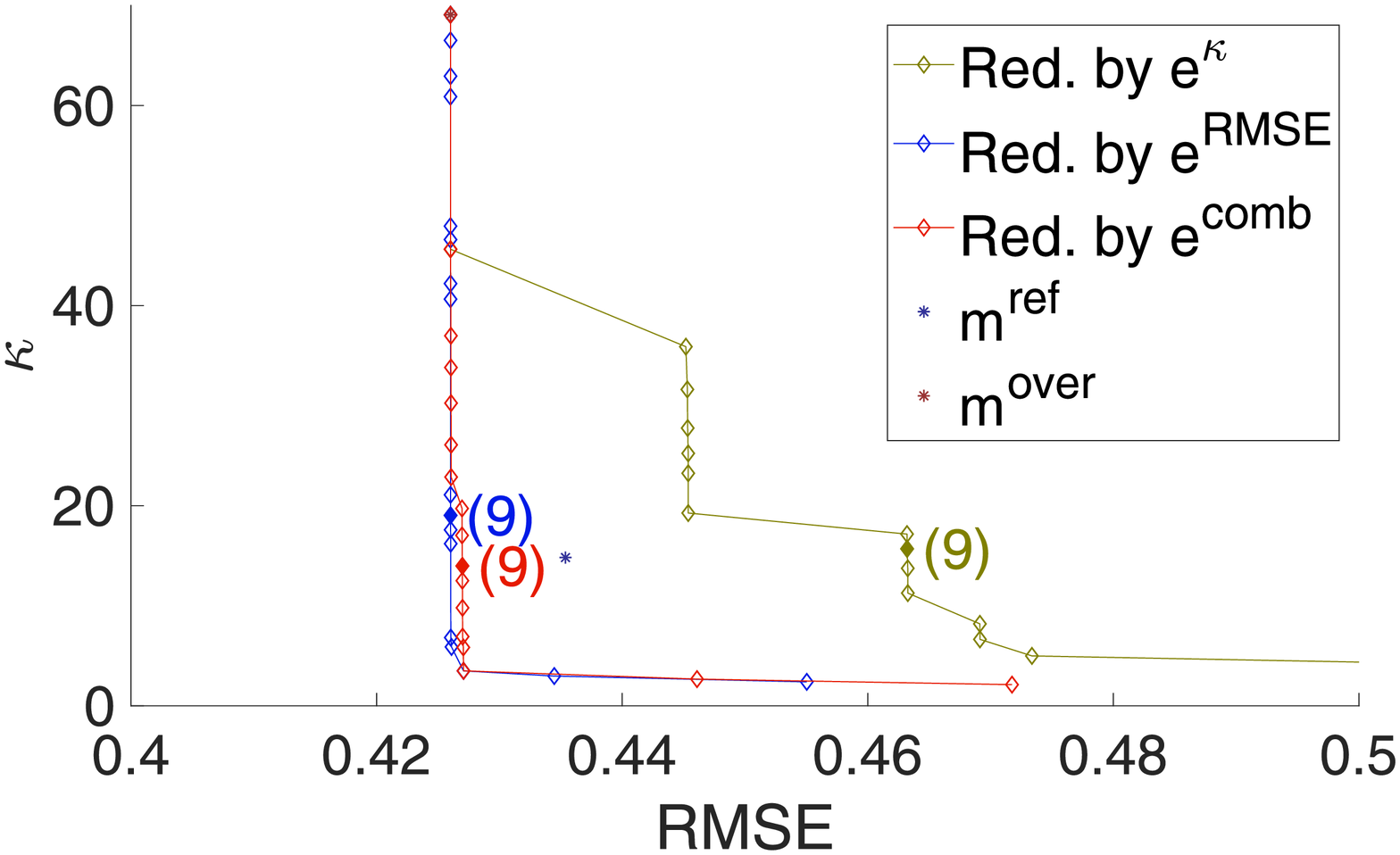}
\label{fig_2a}} &
\subfloat[N-FINDR]{\includegraphics[width=0.26\textwidth,trim=0 12 0 8,clip]{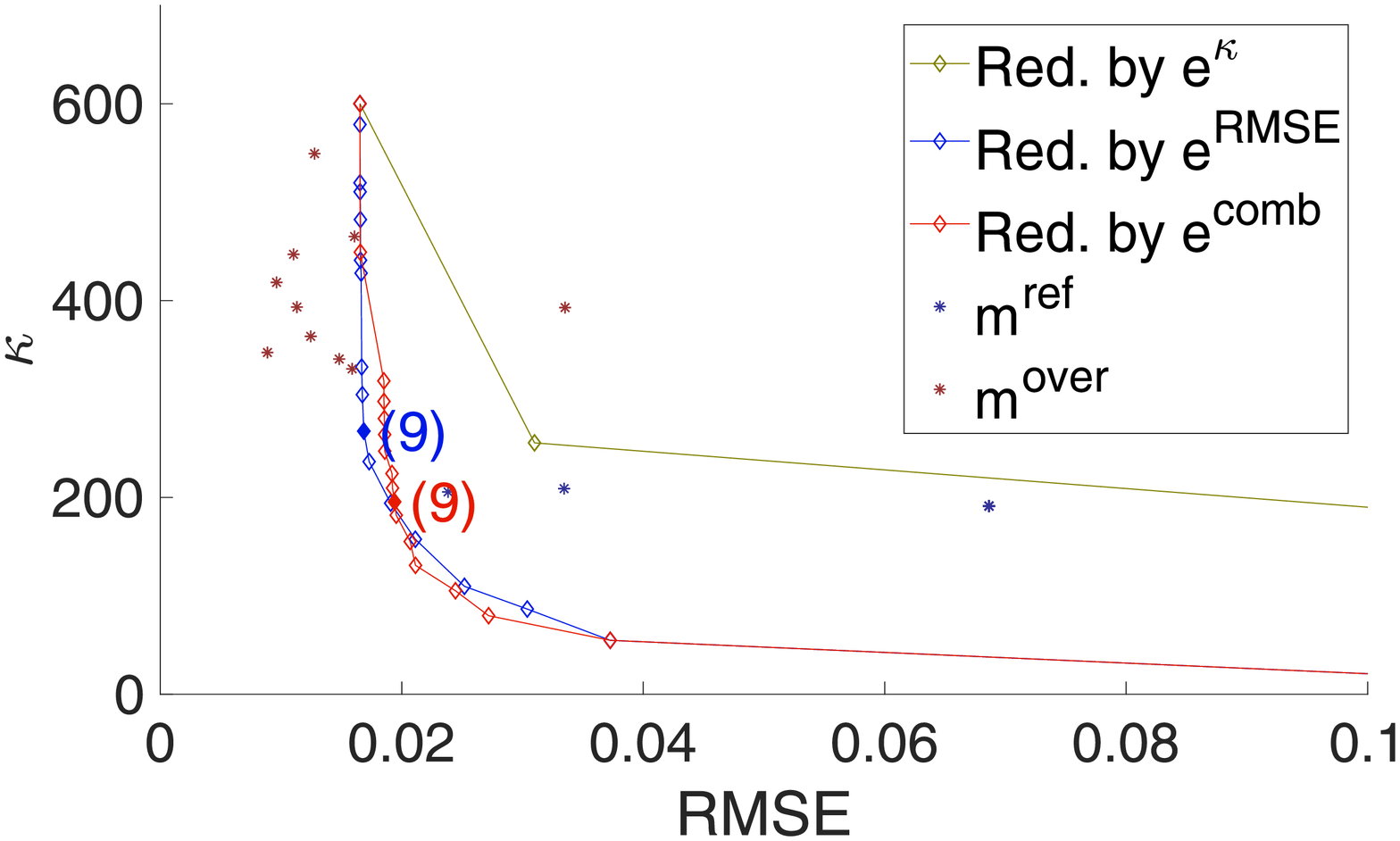}
\label{fig_2b}} &
\subfloat[VCA]{\includegraphics[width=0.26\textwidth,trim=0 14 0 8,clip]{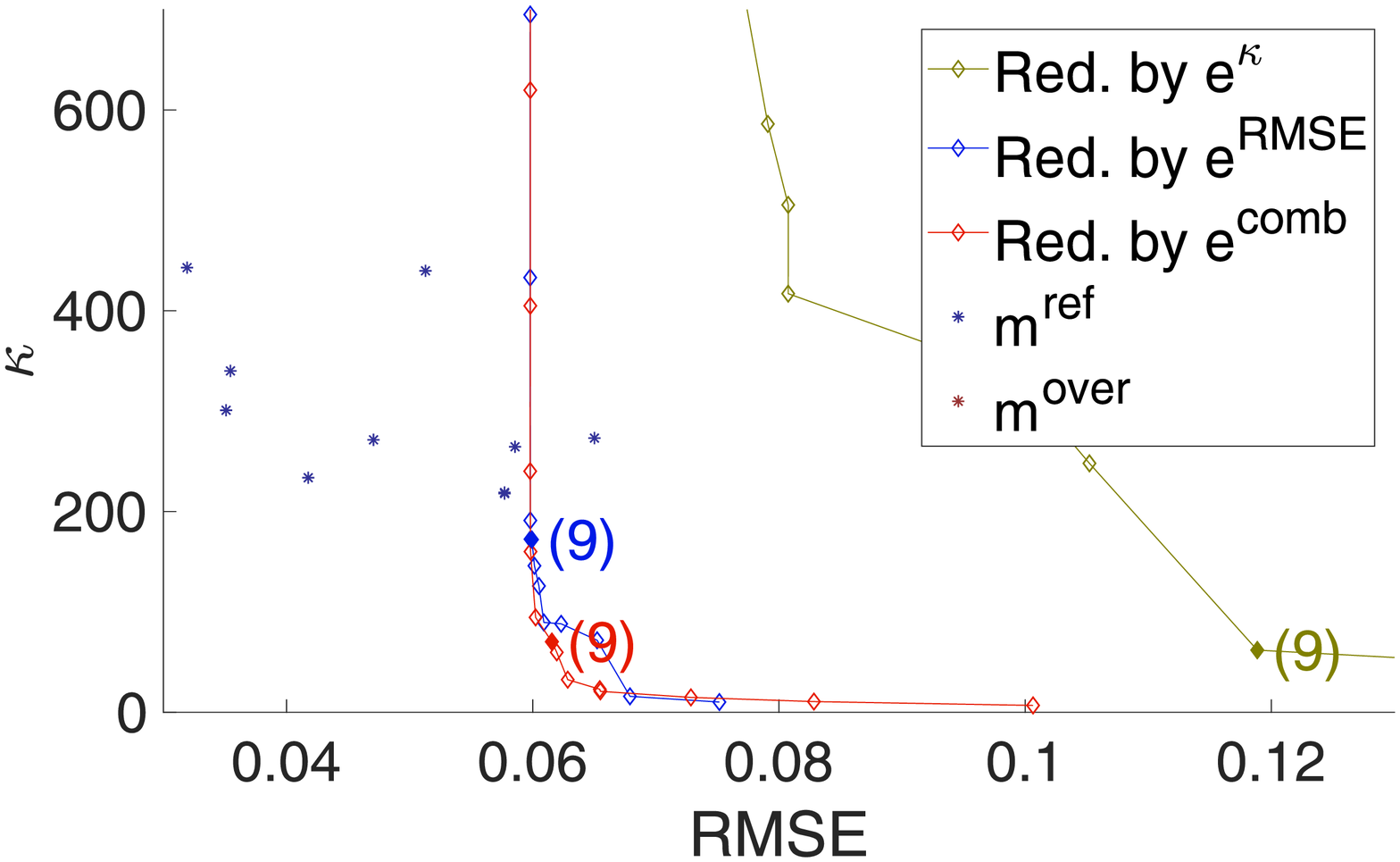}
\label{fig_2c}}\\[-6.5em] 
\rotatebox[origin=c]{90}{\hspace*{24mm} Cuprite} &
\subfloat[OSP]{\includegraphics[width=0.26\textwidth,trim=0 14 0 8,clip]{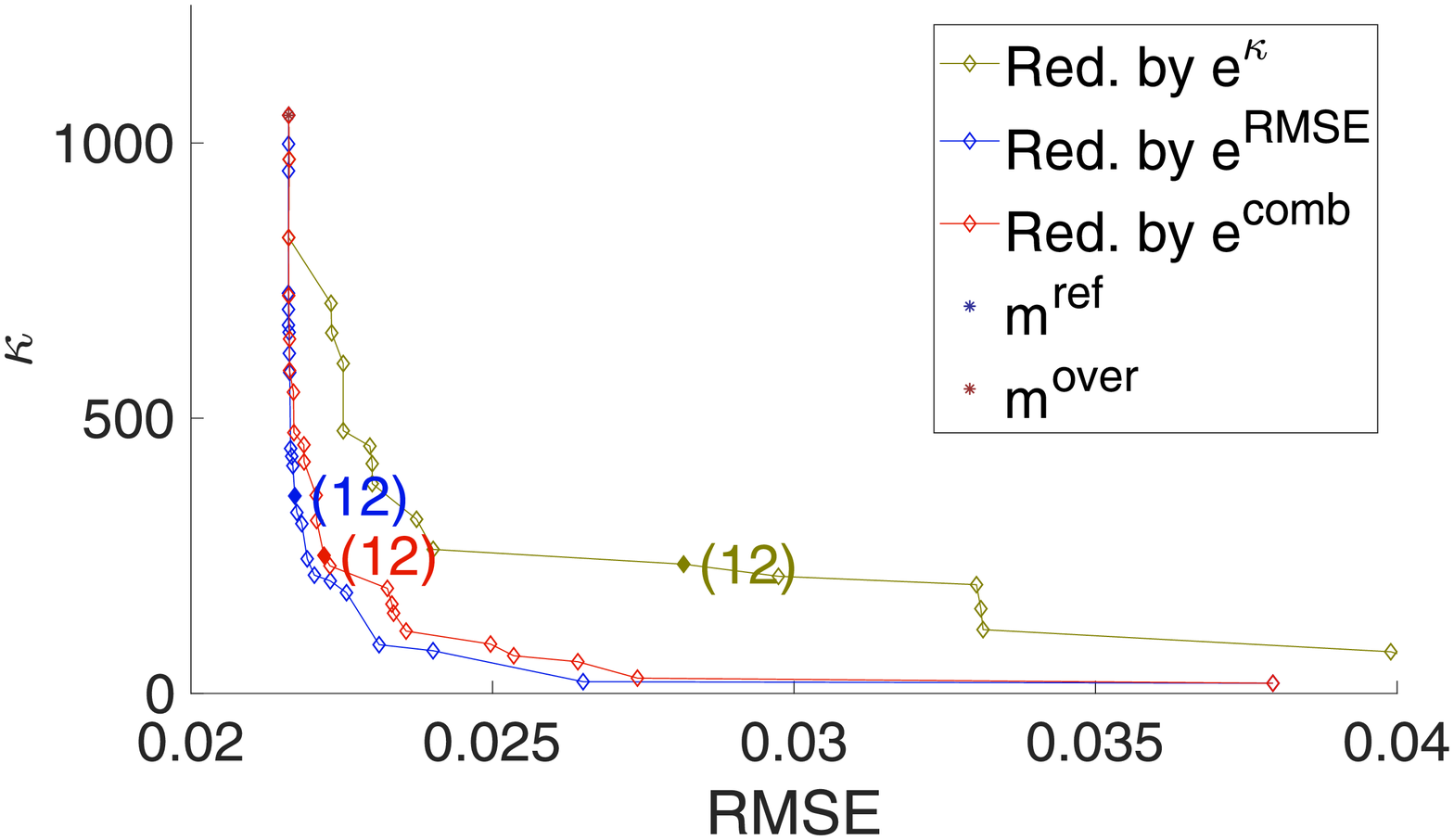}
\label{fig_2d}} &
\subfloat[N-FINDR]{\includegraphics[width=0.26\textwidth,trim=0 14 0 8,clip]{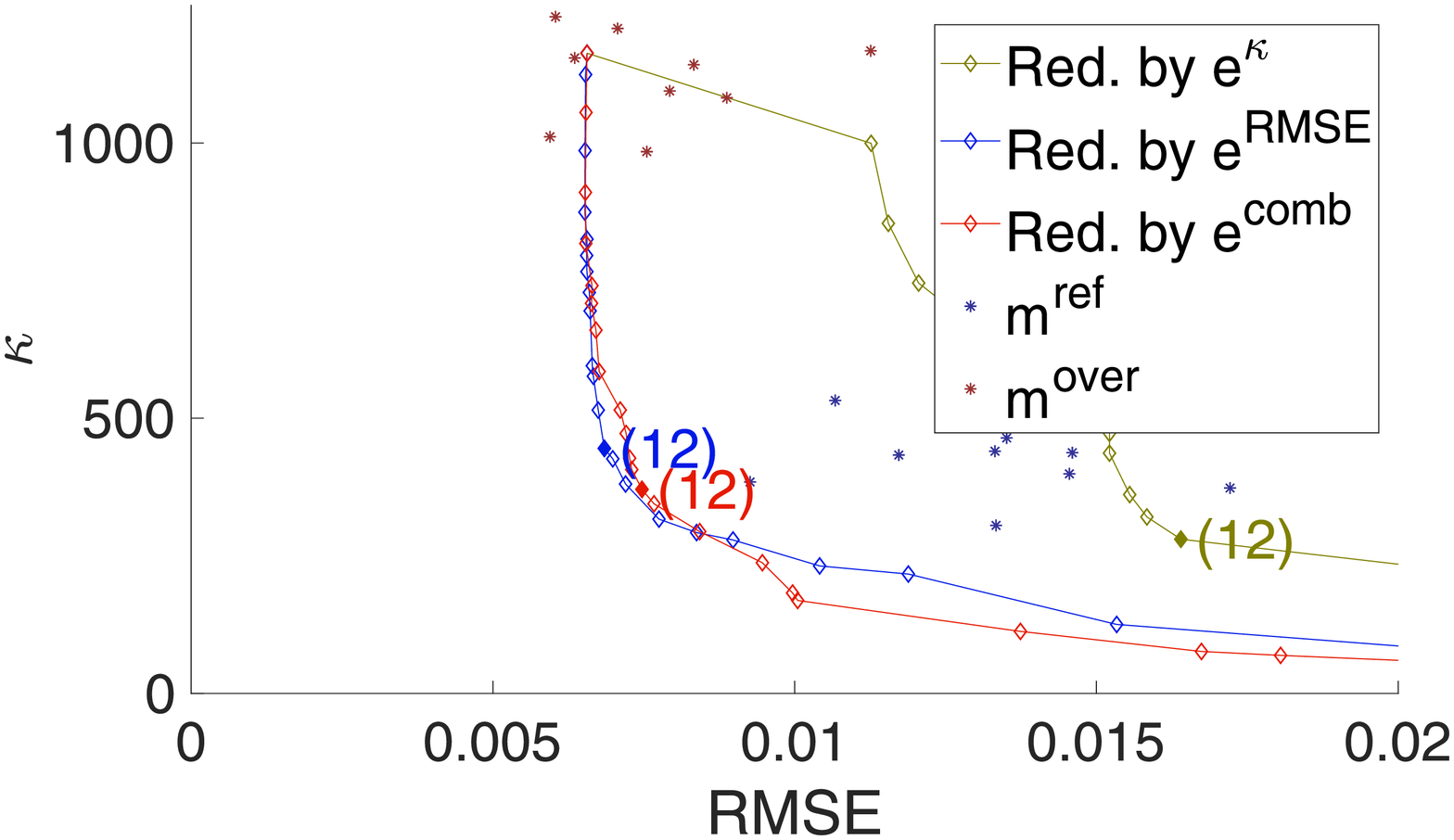}
\label{fig_2e}} &
\subfloat[VCA]{\includegraphics[width=0.26\textwidth,trim=0 14 0 8,clip]{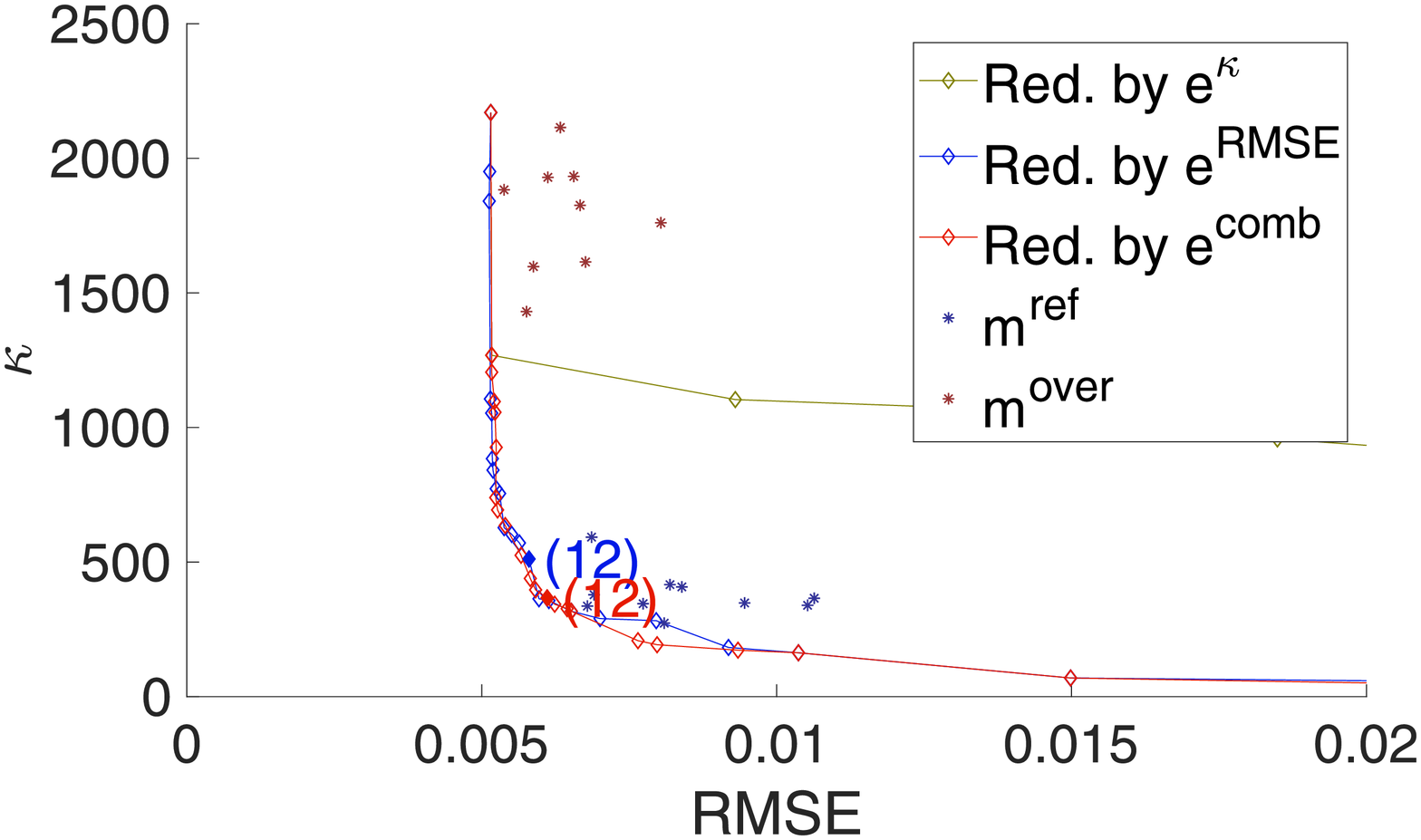}
\label{fig_2f}}\\[-4.5em]
\rotatebox[origin=c]{90}{\hspace*{24mm} Indian Pines} &
\subfloat[OSP]{\includegraphics[width=0.26\textwidth,trim=0 14 0 8,clip]{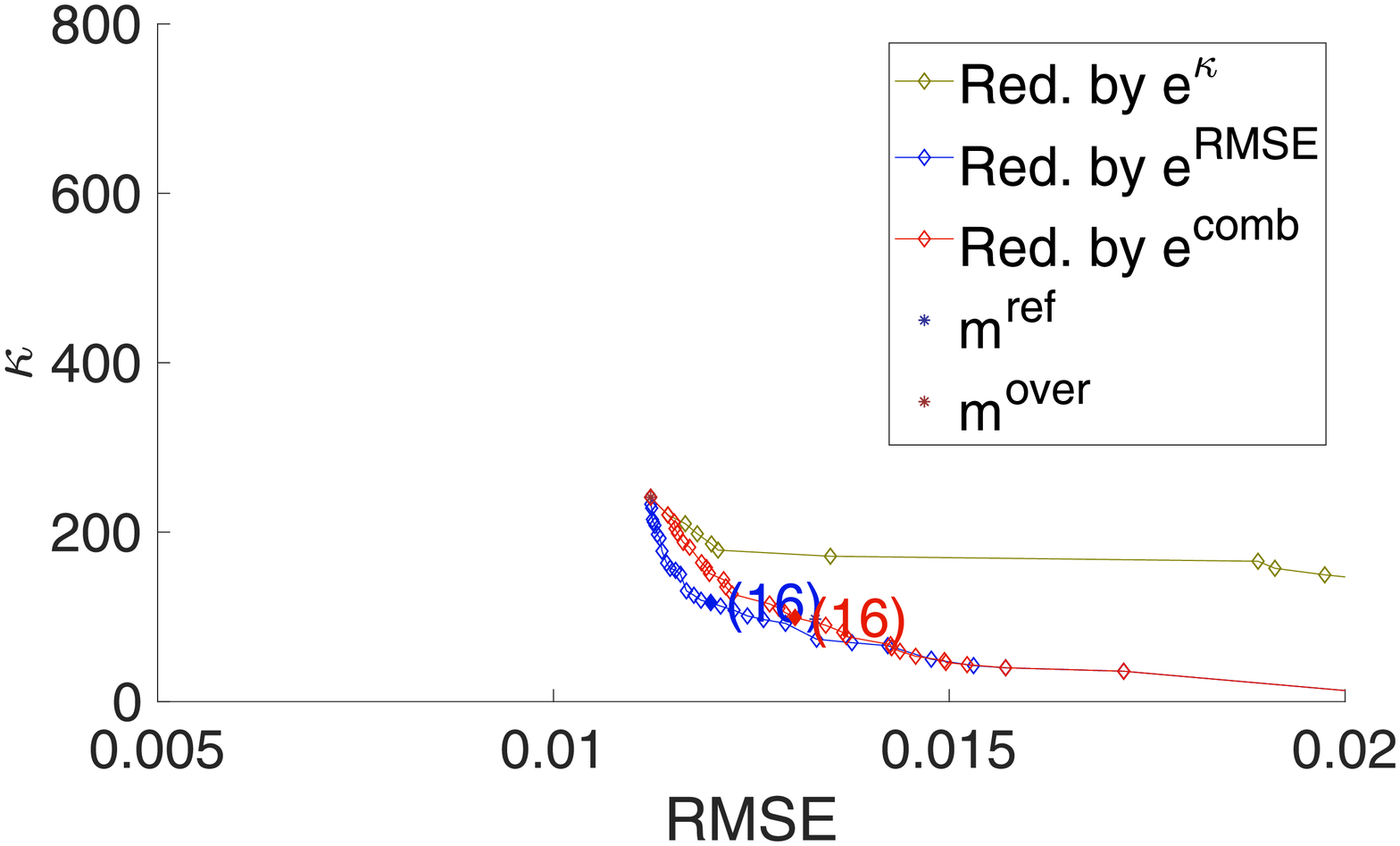}
\label{fig_2g}} &
\subfloat[N-FINDR]{\includegraphics[width=0.26\textwidth,trim=0 14 0 8,clip]{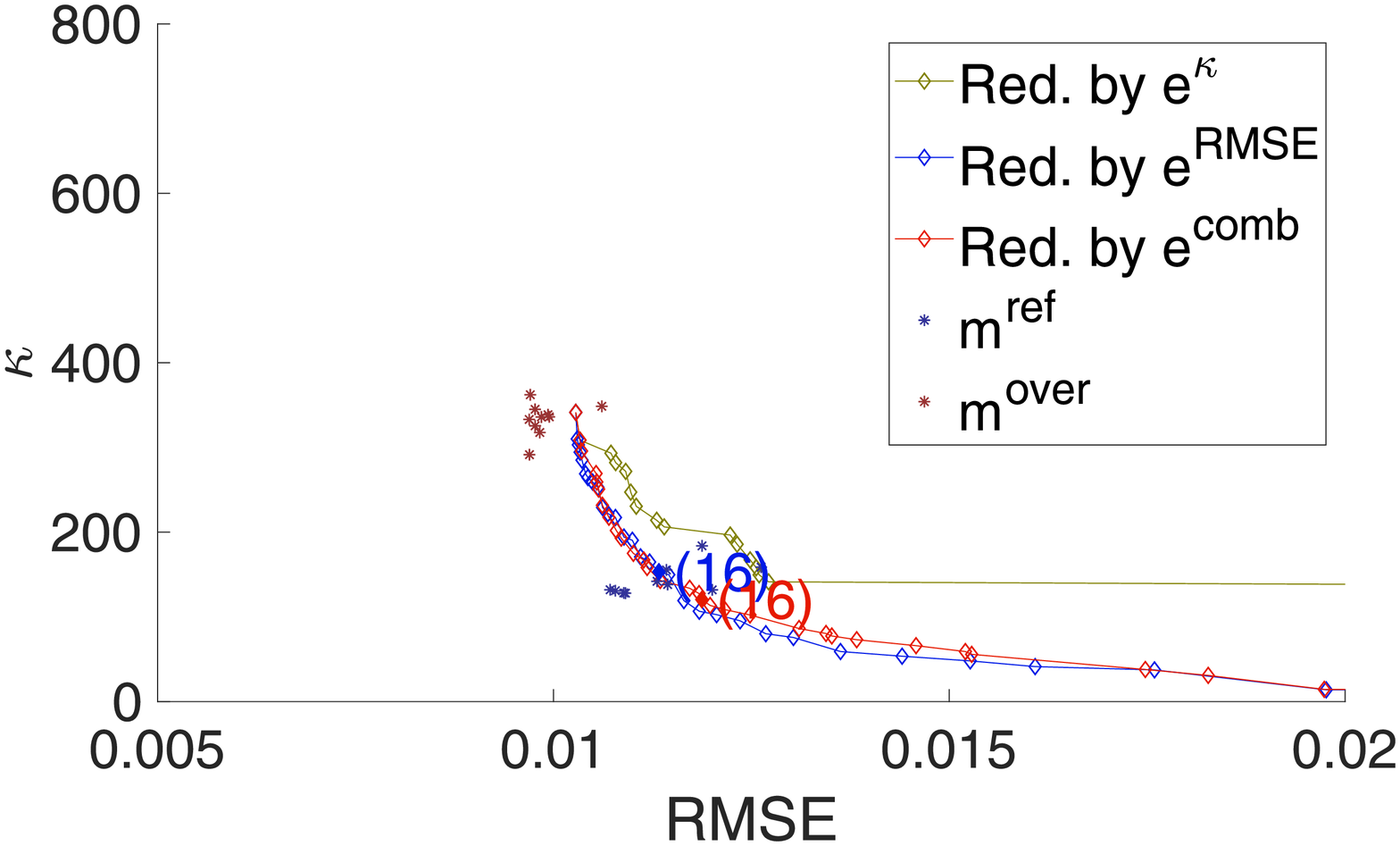}
\label{fig_2h}} &
\subfloat[VCA]{\includegraphics[width=0.26\textwidth,trim=0 14 0 8,clip]{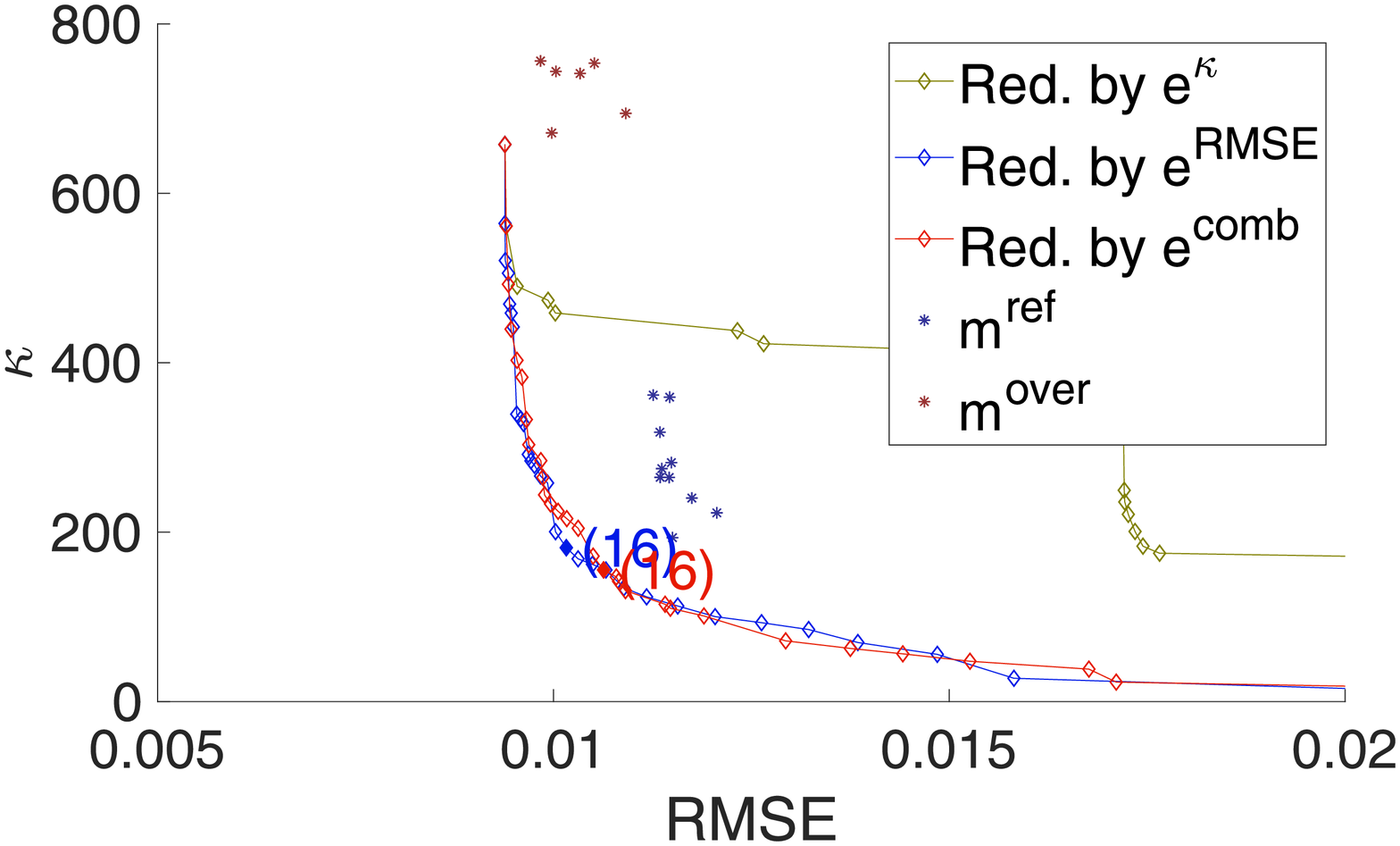}
\label{fig_2i}}\\[-5.5em]
\rotatebox[origin=c]{90}{\hspace*{24mm} Kennedy Space Center} &
\subfloat[OSP]{\includegraphics[width=0.26\textwidth,trim=0 14 0 8,clip]{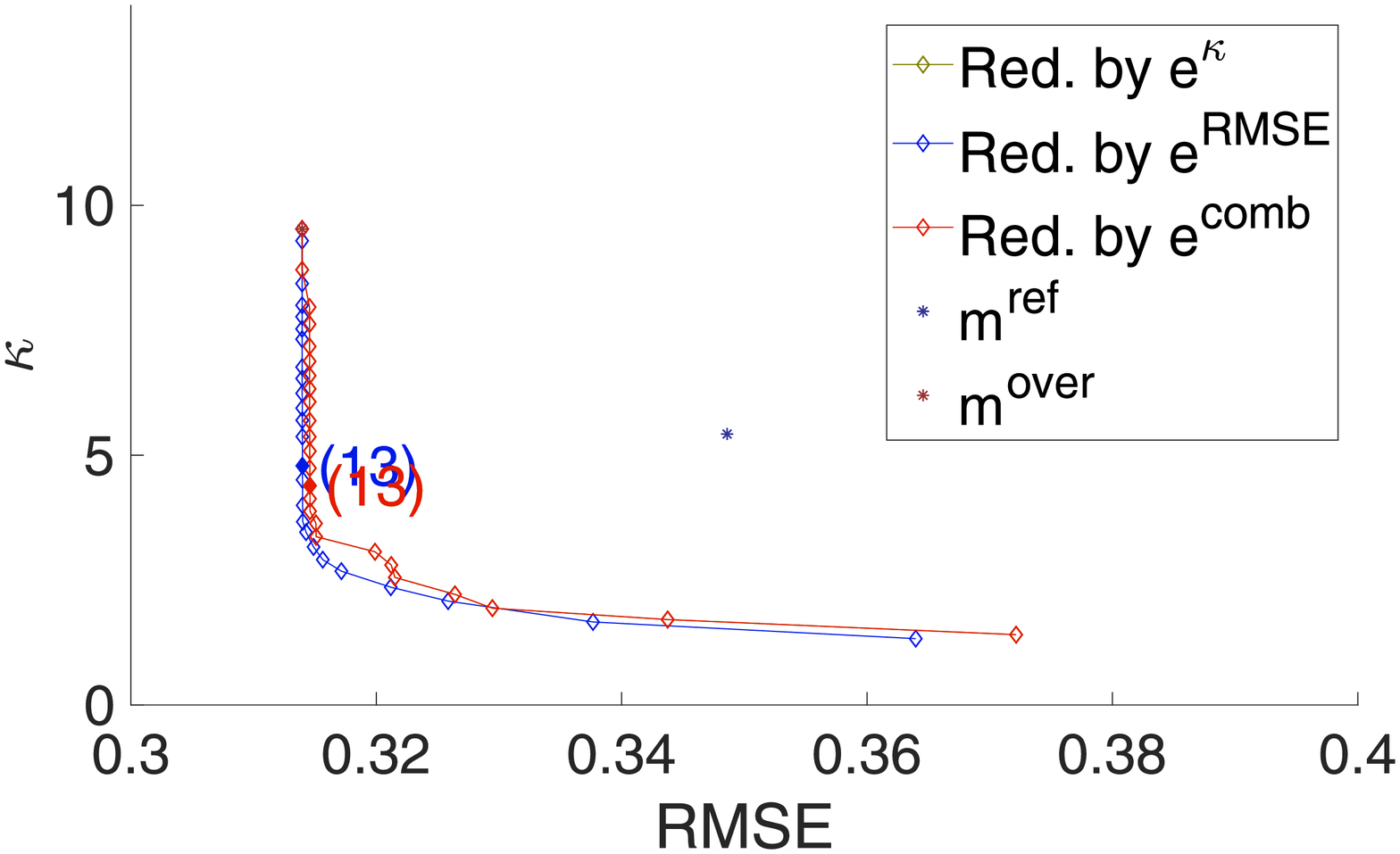}
\label{fig_2j}} &
\subfloat[N-FINDR]{\includegraphics[width=0.26\textwidth,trim=0 14 0 8,clip]{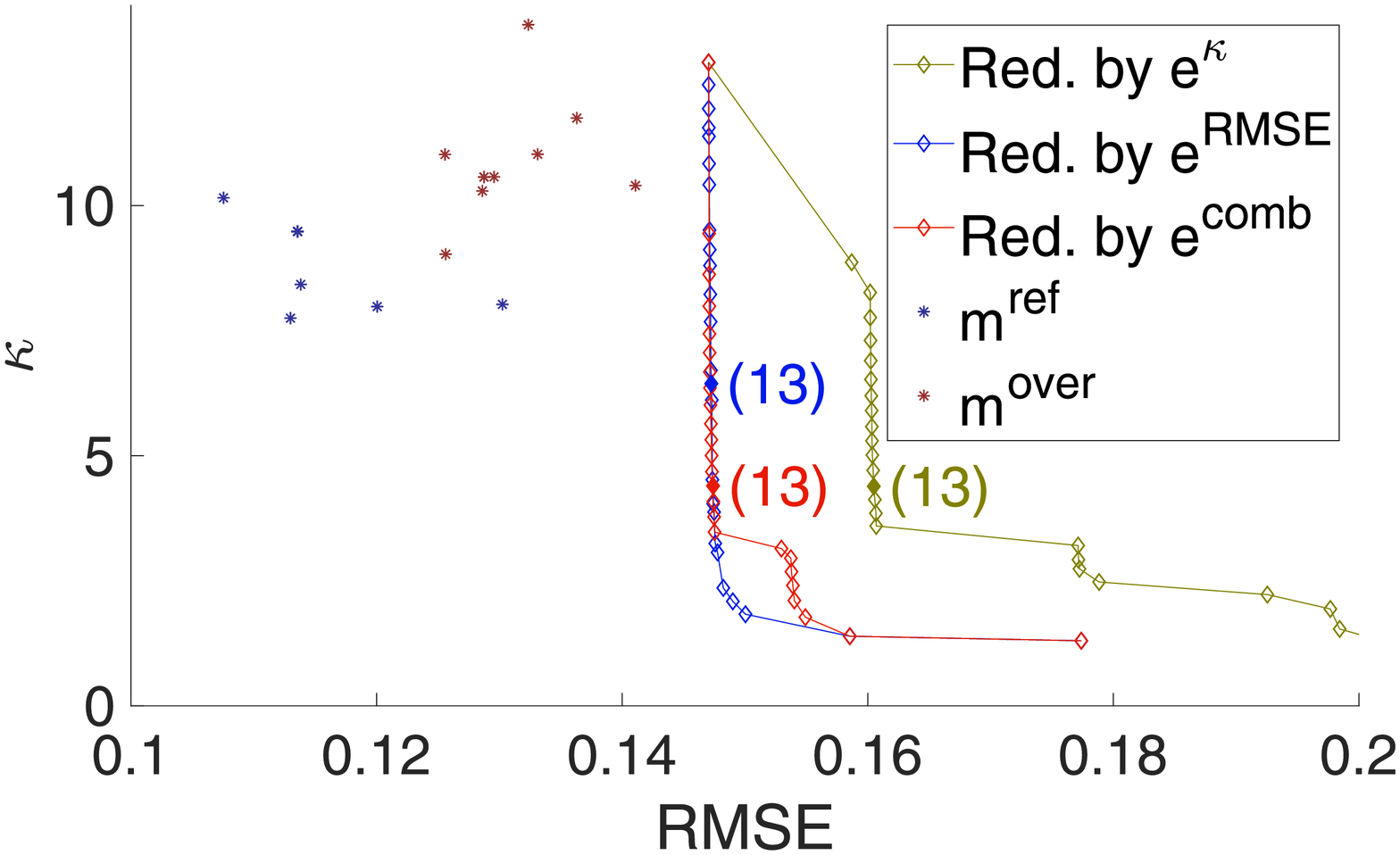}
\label{fig_2k}} &
\subfloat[VCA]{\includegraphics[width=0.26\textwidth,trim=0 14 0 8,clip]{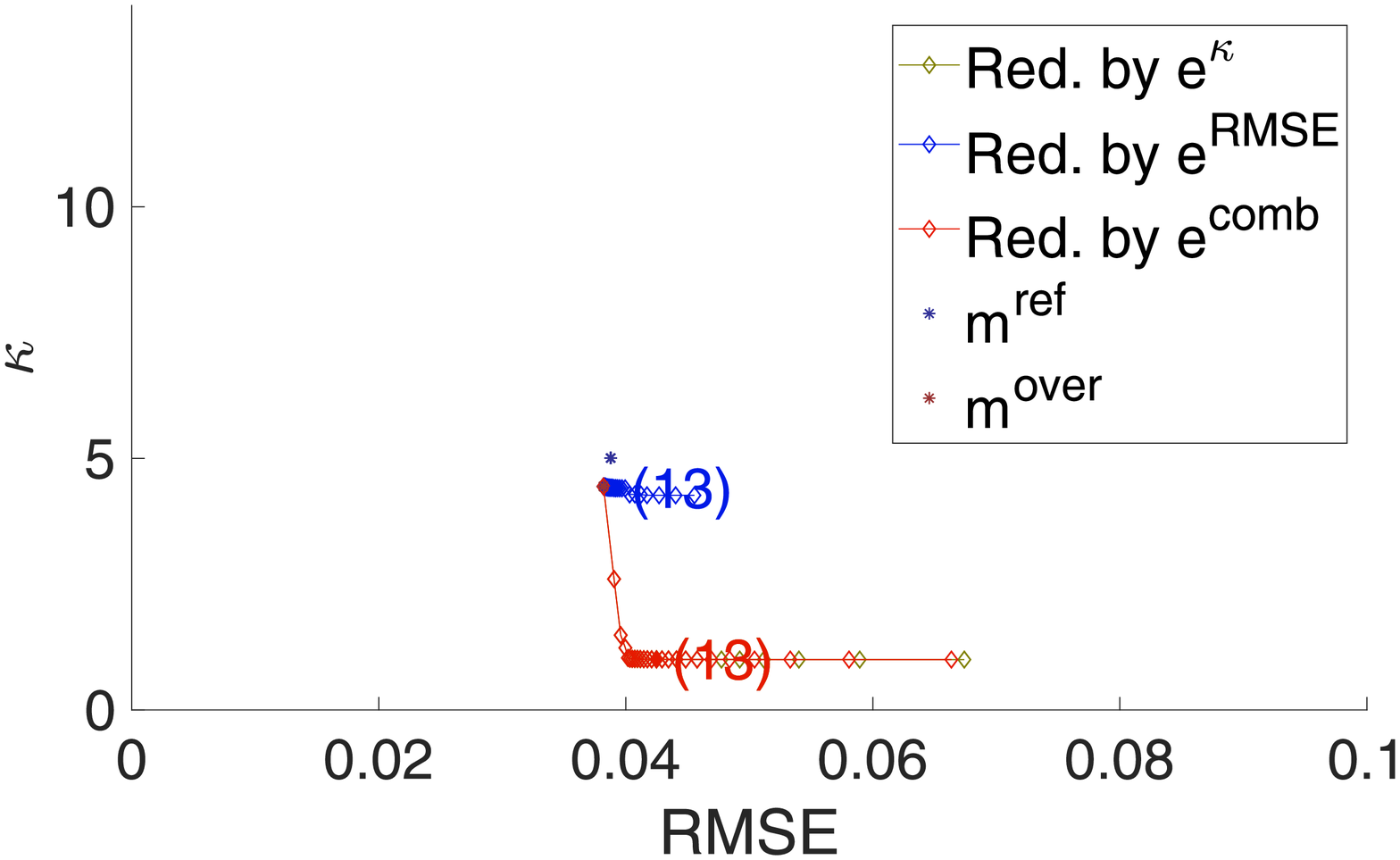}
\label{fig_2l}}
\end{tabular}\\[-7em]
\caption{Results, Pavia University: \eqref{fig_2a} OSP, \eqref{fig_2b} N-FINDR, \eqref{fig_2c} VCA (cropped $m^\text{over}$ EM sets with $\kappa\in[850,2500]$ and $\text{RMSE}\in[0.05,0.07]$); Cuprite: \eqref{fig_2d} OSP (cropped $m^\text{ref}$ with $\text{RMSE}=0.0434$ and $\kappa = 242$), \eqref{fig_2e} N-FINDR, \eqref{fig_2f} VCA; Indian Pines: \eqref{fig_2g} OSP, \eqref{fig_2h} N-FINDR, \eqref{fig_2i} VCA; Kennedy Space Center: \eqref{fig_2j} OSP (Red. by $e^\kappa$ = Red. by $e^{\text{comb}}$), \eqref{fig_2k} N-FINDR, \eqref{fig_2l} VCA (all $m^\text{over}$ and all $m^\text{ref}$ are equal).}
\label{fig_2}
\end{figure*}

\subsection{Evaluation Scheme}
\label{E_CCRD}
To evaluate both, our condition-residuum diagram and our reduction algorithm, we plot different and additional information in the diagram that would not be determined in a practical use-case (see Fig.~\ref{fig_1}). We plot the reduction curves based on the removal of $e^{\kappa} (\alpha=0), e^{\text{comb}} (\alpha=\frac12)$ and $e^{\text{RMSE}} (\alpha=1)$ (see Sec.~\ref{methodB}). 
Additionally, we display EM set resulting from EE algorithms that directly generate $m^{\text{over}}$ and $m^{\text{ref}}$ EMs. For non-deterministic algorithms, \ie N-FINDR and VCA, we generate 10 EM sets, for the deterministic OSP algorithm only one EM set with
$m^{\text{over}}$ and $m^{\text{ref}}$ EMs.
For the Salinas-A dataset in Fig.~\ref{fig_1}, we additionally randomly generate all possible subset of $\mathcal{S}_{m^{\text{over}}}$ containing $m^{\text{ref}}$ EMs denoted as ``Bruteforce''. 

Note, that we crop the diagram to the area close to the ideal condition-residuum point and, thus, discard examples far off the region of good EM sets. Therefore, in some of the diagrams not all direct EM set extractions with $m^{\text{ref}}$ and/or $m^{\text{over}}$ are cropped as well, if there condition-residuum values are out of the area of interest. 

\subsection{Quality of Reduction Schemes}
\label{Red}
\paragraph*{Reduction Schemes}
Considering the general shape of the reduction curves, all of them are nearly L-shaped, where the most interesting region is in the kink of the L, close to the optimal point ($\kappa=1, \text{RMSE}=0$). The $\alpha$-parameter properly controls the impact of both measures on the reduction process, \ie $\kappa$ and RMSE. If the reduction completely relies on $e^\kappa$ ($\alpha=0$), the reduction leads to a fast increase in RMSE, while the condition-number $\kappa$ still stays on a high level. Thus, it is advisable to use higher values of $\alpha\geq\frac12$ leading to a higher influence of the relative RMSE measure in Eq.~\ref{eq:cost-func}. The reduction schemes by $e^{\text{RMSE}}$ ($\alpha=1$) and $e^{\text{comb}}$ ($\alpha=\frac12$) lead to good results. As expected, the reduction by $e^{\text{comb}}$ tends to faster reduce the condition-number at the cost of a slightly increased RMSE. Comparing the full curves, the reduction by $e^{\text{comb}}$ and by $e^{\text{RMSE}}$ are alike. While the $e^{\text{comb}}$-reduction curve mainly runs below the reduction by $e^{\text{RMSE}}$ in Figs~\ref{fig_1a}, \ref{fig_1b}, \ref{fig_2c}, \ref{fig_2e}, \ref{fig_2f}, and \ref{fig_2l}, the opposite is the case in Figs.~\ref{fig_2d}, \ref{fig_2g}, \ref{fig_2h}, \ref{fig_2j}, and \ref{fig_2k}. Also, when considering the kink of L-shaped curves, both methods can delivers more regular shapes, \ie, reducing by $e^{\text{comb}}$ delivers superior results in Figs.~\ref{fig_2c}, \ref{fig_2f}, and \ref{fig_2l},whereas reduction by $e^{\text{RMSE}}$ is better inFigs.~\ref{fig_2j}, and \ref{fig_2k}.

\paragraph*{Bruteforce Results}
Fig.~\ref{fig_1} shows all random subsets of $\mathcal{S}_{m^{\text{over}}}$ with $m^{\text{ref}}$ EMs. Obviously, our reduction schemes based on the reduction of $e^{\text{comb}}$ and $e^{\text{RMSE}}$ select EM sets with as good as possible condition-number and RMSE among all possible EM subsets.

\paragraph*{Direct EM set Extraction}
Evaluating of our method versus direct EE algorithm that generate $m^{\text{ref}}$ EMs, the combined reduction scheme (removing $e^{\text{comb}}$) and the RMSE reduction (removing $e^{\text{RMSE}}$) commonly exhibit good results. This is due to the fluctuation of these methods in generating the initial over-complete EM set. Considering the reduction by $e^{\text{comb}}$ with direct EE results in EM sets with better condition-number and RMSE (Figs.~\ref{fig_2a}, \ref{fig_2b}, \ref{fig_2e}, \ref{fig_2f}, \ref{fig_2i}, \ref{fig_2j}), with better RMSE but worse condition-number (Figs.~\ref{fig_1a}, \ref{fig_1b}, \ref{fig_1c}, \ref{fig_2d}, \ref{fig_2g}) or with better condition-number but worse RMSE (Figs.~\ref{fig_2c}, \ref{fig_2h},\ref{fig_2k}, \ref{fig_2l}). Most of the latter two cases are due to the non-deterministic nature of the underlying EE algorithms, \ie N-FINDR and VCA. See also Sec.~\ref{limitations}, where we discuss the specific situation in Figs.~\ref{fig_2c} and \ref{fig_2k}.

\section{Discussion}
\subsection{Optimal Size of Endmember-Sets}
\label{ONoE}
Our residuum-condition diagram provides a visual guidance in selecting EM sets with low RMSE and low condition number (high unmixing stability), \ie, EM sets that are close to the optimal case (RMSE$=0$ and $\kappa=1$). In all of our text cases, the reduction curves based on the removal of $e^{\text{RMSE}}$ and $e^{\text{comb}}$ result in are quite pronounced shape that indicates EM sets close to the theoretic optimum. Compared to the reference EM set sizes $m^{\text{ref}}$ provided in literature, we see, that in some cases these reference sizes are located close to the main bend of the curve (see Fig.~\ref{fig_1c}, \ref{fig_2b}, \ref{fig_2c}, \ref{fig_2d},\ref{fig_2f}), while in other cases the reference EM set sizes are too conservative, \ie, smaller EM sets lead to more stable results at minimal loss in RMSE (see Fig.~\ref{fig_2a}, \ref{fig_2j}, \ref{fig_2k}), or too progressive, \ie larger EM sets lead to significant lower RMSE at minimal loss in stability (see Fig.~\ref{fig_1a}, \ref{fig_1b}, \ref{fig_2l}). In general, our semi-automatic EM set selection approach easily supports the choice of the EM set size (and potentially the EM set itself) from an application perspective. It can easily be used in combination with automatic EM set size estimation algorithms~\cite{Asl2016}.

\subsection{Algorithmic Complexity}
\label{AlgoComplexity}
Our reduction approach starting with an over-complete EM set requires $\frac{m(m+1)}2$ unmixing steps at each level, where $m$ is the size of the current EM set (brute force testing would be of exponential order). As fully constrained unmixing is computationally quite exhaustive, we experimented using unconstrained unmixing, which is computationally far less demanding. In approximately 50\% of our tests, the results have been qualitatively the same as with fully constrained unmixing, \ie the reduction curve's shape and relative location of the EM sets on both curves are very close. In the rest of the cases, both reduction curves significantly differ from each other, thus selecting the set size from the unconstrained unmixing may lead to wrong interpretations in these cases. Making our reduction scheme more efficient is part of our future research.

\subsection{Limitations}
\label{limitations}
The result of our reduction approach strongly depends on the quality of the initial over-complete EM set $\mathcal{S}_{m^{\text{over}}}$. For non-deterministic EE algorithms, the spread of the initial EM sets may be quite significant; (\eg Fig.~\ref{fig_2b}). Two very specific cases are the N-FINDR result applied to the Kennedy-Space-Center dataset (Fig.~\ref{fig_2k}) and the VCA result applied to the Pavia University dataset (Fig.~\ref{fig_2c}). Here, the reduction schemes deliver significantly worse results then directly extracted EM set with the ``optimal'' set size $m^{\text{ref}}$. These results are quite counter-intuitive, as N-FINDR and VCA deliver worse results in terms of RMSE using a $m^{\text{over}}=2\cdot m^{\text{ref}}$ compared to $m^{\text{ref}}$ EMs. Thus, it may be advisable to run any EE algorithm \emph{after} having semi-automatically selected the EM set size using our reduction scheme.

\subsection{EE Algorithms Comparison}
\label{Algo}
Even though it is not the goal of our paper to explicitly compare EE algorithms, our evaluation implies some tendencies. In some cases OSP delivers high quality results (e.g. Fig.~\ref{fig_1a}), but in most cases its reconstruction quality falls behind that of N-FINDR and VCA. When directly generating EM sets, the spread of the N-FINDR results are less compared to VCA (see \eg Figs~\ref{fig_2h} and \ref{fig_2h}). Factoring out the spread of both, N-FINDR and VCA, these EE algorithms show a comparable performance. The evaluation shows every EE algorithm clearly benefits from our reduction scheme.

\section{Conclusion}
\label{conclusion}
We introduced and analyzed the concept of condition-residuum diagrams in combination with an EM set reduction scheme based on combined condition and residuum optimization, that is applied to over-complete EM set. We show, that this approach can be used as visual guidance in selecting the EM set size and the EM set itself. We evaluated our approach for three common EE algorithms (OSP, N-FINDR and VCA) and with three different energy functionals for optimized EM reduction. Here, the RMSE-based and the combined RMSE-condition schemes show good results with a slight advantage in favor of the combined scheme. In the future, we will investigate data-driven approaches to steer the mixture parameter $\alpha$, alternative optimization energy functionals and EM replacement approaches, as well as means of accelerating the costly full-constrained unmixing during reduction.
Furthermore, combining our reduction approach with a spectral feature selection or spectral weighting approach might be beneficial.

\section*{Acknowledgment}

This work was supported by the German Research Foundation (DFG) under research grant KO~2960/10-2.

\ifCLASSOPTIONcaptionsoff
  \newpage
\fi



\bibliographystyle{IEEEtran}


\bibliography{paper}


%
%

\end{document}